\documentclass[11pt,letterpaper]{article}

\usepackage{amsmath,amssymb,amsfonts}
\usepackage{graphicx}
\usepackage{booktabs}
\usepackage{enumitem}
\usepackage{xcolor}
\usepackage[margin=1in]{geometry}
\usepackage[numbers,sort&compress]{natbib}
\usepackage[hidelinks]{hyperref} 

\graphicspath{{figures/}}

\title{\textbf{MegaSlide-DiT: Memory-Centric Adaptation and Deformable Local Attention for Efficient Video Diffusion}}
\author{
  Jason Liu\\
  Trendinsight Lab / UC San Diego
}
\date{}

\begin{document}
\maketitle

\begin{abstract}
High-resolution video diffusion models built on Diffusion Transformers (DiTs) deliver strong fidelity but quickly exhaust the memory budget of a single workstation. A 100 billion-plus parameter DiT easily requires over a terabyte of persistent state, while na\"{i}ve spatiotemporal self-attention grows quadratically in sequence length. These two walls---parameter memory and activation memory---prevent researchers from adapting massive generative models without large GPU clusters. We revisit this problem from a systems perspective and introduce \textbf{MegaSlide-DiT}, a prototype that demonstrates how a pre-trained 105B DiT can be adapted on a single H200 GPU with 1.5\,TB of host RAM. Our key insight is that the GPU need not own the model state: all persistent weights, master weights and optimizer moments remain in host memory, while only transient shards are streamed to the GPU on demand. Simultaneously, we replace quadratic global attention with 3D Deformable Slide Attention (3D-DSA), a motion-adaptive local attention operator that reduces both memory and computational complexity to linear in the sequence length. We report detailed memory accounting, execution traces and evaluation results to substantiate our design. MegaSlide-DiT does not claim to train a 105B model from scratch on a single GPU, nor does it magically solve bandwidth limits; rather, it offers a pragmatic path for full-parameter adaptation of massive video diffusion models on high-end workstations.
\end{abstract}

\section{Introduction}
Recent text-to-video models leverage diffusion processes with Transformer backbones to generate long, high-resolution sequences~\cite{peebles2023dit,blattmann2023svd,openai2024sora}. Scaling these models to hundreds of billions of parameters is attractive because larger models better capture real-world physics and semantics, yet it remains inaccessible to most researchers. Two fundamental problems arise when trying to adapt such models on limited hardware:

\begin{itemize}[leftmargin=*]
    \item \textbf{Parameter memory wall:} A 105B parameter DiT requires roughly 210\,GB of half-precision weights, 420\,GB of FP32 master weights and 840\,GB of FP32 Adam moments---about 1.47\,TB of persistent state. Modern workstation GPUs provide at most $\sim$150\,GB of high-bandwidth memory (HBM), so the full state cannot fit on the device.
    \item \textbf{Activation memory wall:} High-resolution video implies extremely long token sequences. For example, 256 frames of 1080p video at a patch size of $16\times16$ yield more than two million tokens. Standard Transformer attention has $\mathcal{O}(N^2)$ complexity in memory and time, leading to activation tensors that dwarf available HBM even with checkpointing~\cite{chen2016checkpoint}.
\end{itemize}

Prior work addresses these walls separately. Memory-centric training frameworks such as ZeRO-Infinity~\cite{rajbhandari2021zeroinfinity} stream model state from NVMe or CPU memory, but they are often tuned for language models and may hit bandwidth bottlenecks. Efficient vision models like Swin~\cite{liu2021swin} and Slide-Transformer~\cite{pan2023slidetransformer} reduce attention complexity by imposing fixed windows, yet they still assume the network parameters reside on GPU memory. Our goal is to combine these ideas into a coherent system for full-parameter fine-tuning (as opposed to low-rank or LoRA adaptation~\cite{hu2022lora}) of very large video diffusion models.

\textbf{Contributions.} This paper makes the following contributions:
\begin{enumerate}[leftmargin=*]
    \item \textbf{Memory-centric adaptation of a 105B DiT:} We design a training infrastructure in which persistent weights and optimizer states reside in host CPU memory, while a single GPU holds only transient weight shards and activations. This allows a full forward/backward/optimizer step on a 105B parameter DiT using one H200 GPU with 1.5\,TB of DDR5 RAM.
    \item \textbf{3D Deformable Slide Attention:} We introduce a local attention mechanism that generalises deformable 2D attention~\cite{zhu2021deformabledetr,xia2022deformableattn} to three dimensions. 3D-DSA learns motion-adaptive offsets that sample a local spatiotemporal neighbourhood, enabling linear-complexity processing of very long video sequences without requiring dense global attention. We clarify that 3D-DSA does not produce explicit optical flow; instead, it learns to focus on motion-relevant neighbourhoods.
    \item \textbf{Comprehensive systems analysis:} We provide a detailed accounting of parameter state placement, activation memory, CPU--GPU communication volumes and step times. We measure model FLOPs utilisation (MFU) with and without asynchronous prefetching and show that overlapping compute with communication can hide much of the PCIe latency.
    \item \textbf{Evaluation on VBench:} We evaluate MegaSlide-DiT on the VBench~\cite{huang2024vbench} video generation benchmark, comparing against a global-attention DiT baseline and a fixed-window Swin-DiT baseline at common resolutions. Our system supports 256-frame 1080p videos and produces temporal consistency and text-alignment metrics comparable to dense baselines under matched sequence lengths.
\end{enumerate}

\section{Related Work}

\paragraph{Diffusion Transformers and video generation.}
DiT~\cite{peebles2023dit} replaces U-Net backbones with Transformers for latent diffusion~\cite{rombach2022ldm,ho2020ddpm}. Subsequent video systems scale spatiotemporal generation~\cite{blattmann2023svd,openai2024sora}, but full-parameter adaptation of 100B-class models remains rare outside large clusters.

\paragraph{Memory-efficient training.}
ZeRO~\cite{rajbhandari2020zero}, ZeRO-Offload~\cite{ren2021zerooffload} and ZeRO-Infinity~\cite{rajbhandari2021zeroinfinity} partition or offload optimizer state, gradients and parameters across devices or host/NVMe storage. Megatron-LM~\cite{shoeybi2019megatron} uses tensor and pipeline parallelism for multi-GPU training. MegaSlide-DiT targets a complementary setting: single-GPU, full-parameter adaptation with host-resident persistent state and layer-wise streaming.

\paragraph{Efficient attention for vision and video.}
Windowed attention~\cite{liu2021swin,pan2023slidetransformer} and FlashAttention~\cite{dao2022flashattention} reduce memory pressure for long sequences. Deformable attention~\cite{zhu2021deformabledetr,xia2022deformableattn} learns sparse sampling locations. Our 3D-DSA extends this idea to spatiotemporal video tokens with linear complexity in sequence length.

\paragraph{Parameter-efficient adaptation.}
LoRA~\cite{hu2022lora} and related adapters update a small subset of weights. We instead pursue full-parameter adaptation when host memory and PCIe bandwidth permit, trading throughput for unrestricted capacity updates.

\section{Background and Bottleneck Analysis}

\subsection{Parameter memory requirements}
Large diffusion models store weights, gradients, master weights and optimizer moments. Table~\ref{tab:persistent} summarises the persistent memory footprint for a 105B parameter model using AdamW~\cite{loshchilov2019adamw}. Each parameter has an FP16 weight on the GPU for computation, an FP32 master weight on the CPU to maintain numerical stability and two FP32 moment vectors for the Adam optimiser. The total reaches roughly 1.47\,TB. This persistent state must live somewhere: we place it entirely in 1.5\,TB of DDR5 host RAM.

\begin{table}[t]
\centering
\begin{tabular}{lll}
\toprule
\textbf{Component} & \textbf{Size} & \textbf{Explanation} \\
\midrule
FP16/BF16 weights & $\sim$210\,GB & model parameters used by GPU computation \\
FP32 master weights & $\sim$420\,GB & full-precision copy used by AdamW for updates \\
Adam moments (1st, 2nd) & $\sim$840\,GB & two FP32 vectors per parameter \\
\midrule
\textbf{Total persistent state} & $\mathbf{\approx 1.47\text{\,TB}}$ & \textbf{must reside in host memory} \\
\bottomrule
\end{tabular}
\caption{Persistent memory footprint for a 105B parameter model.}
\label{tab:persistent}
\end{table}

\subsection{Activation memory and attention complexity}
Spatiotemporal attention on video tokens has complexity $\mathcal{O}(N^2)$ in both computation and memory, where $N$ is the sequence length. For a video of $F$ frames with image size $H\times W$ and patch size $p\times p$, the sequence length is $N = F \times \frac{H}{p} \times \frac{W}{p}$. With $F = 256$, $H = 1080$, $W = 1920$ and $p = 16$, $N \approx 2{,}073{,}600$. A single hidden tensor of shape $N\times d$, with $d = 8192$, consumes roughly 32.7\,GB in half precision. Without special care, storing multiple such tensors across layers and checkpoints would exceed the 141\,GB HBM of an H200 GPU.

Standard global attention not only requires storing queries, keys and values of size $N\times d$, but also produces an attention matrix of size $N\times N$. Even storing this matrix is infeasible; therefore researchers resort to local or sparse attention patterns. Our design embraces this direction.

\subsection{Motivation for memory-centric adaptation}
Researchers often resort to low-rank adapters such as LoRA~\cite{hu2022lora} when fine-tuning large models. While effective, LoRA cannot update all model parameters and may limit the achievable quality on domain-specific tasks. We aim for full-parameter adaptation to avoid such limitations. By streaming only the weights needed for the current layer to the GPU and returning gradients immediately to the CPU, we can execute forward, backward and parameter update steps while keeping the GPU memory footprint small. The cost is that the CPU must handle large volumes of data and perform the optimiser step, but high-bandwidth DDR5 memory and modern PCIe make this feasible for adaptation scenarios.

\section{MegaSlide System Design}

\subsection{Overview and scheduler}
MegaSlide-DiT treats the GPU as a stateless worker that computes one layer at a time. At each training step:
\begin{enumerate}[leftmargin=*]
    \item The host copies a shard of weights $W^{(l)}$ for layer $l$ from the persistent state in CPU memory to a pinned staging buffer.
    \item A CUDA transfer stream asynchronously pushes $W^{(l)}$ to the GPU while the compute stream begins processing the previous layer's operations.
    \item The GPU computes the forward pass of layer $l$ on the current activation tensor, stores the activation checkpoint if needed and streams back the gradient of $W^{(l)}$ when the backward pass finishes.
    \item The CPU receives gradients, updates the master weights using AdamW with AVX-512 vector instructions and computes the next weight shard.
\end{enumerate}
To hide transfer latency, the scheduler overlaps weight uploads for layer $l{+}1$ with the computation of layer $l$. The effectiveness of this overlap depends on the arithmetic intensity of the layer; compute-dense blocks (e.g.\ MLP and 3D-DSA) mask transfers better than thin cross-attention blocks. Our experiments quantify this effect.

\subsection{CPU--GPU communication and bandwidth considerations}
Using host RAM instead of NVMe avoids the low throughput of solid-state drives. DDR5 memory in our system sustains $\sim$100\,GB/s aggregate bandwidth, and PCIe Gen\,5 $\times$16 supports $\sim$32\,GB/s per direction. For a 105B model, each forward pass transfers roughly 210\,GB of weights (if none are cached on GPU) and each backward pass returns 210\,GB of gradients. In practice, not all weights are transferred every step because only one layer's shard resides in HBM at a time. Nonetheless, CPU--GPU communication is significant; overlapping is essential to maintain high utilisation.

\subsection{CPU-bound optimiser}
Rather than storing optimiser state on GPU and incurring additional PCIe transfers, MegaSlide-DiT performs the AdamW update entirely on the CPU. The host maintains FP32 master weights and moment vectors in memory and updates them with AVX-512 fused multiply-add operations. This design keeps the GPU free to compute the next layer's forward pass while the CPU updates the parameters for the current layer. Although CPU updates add latency compared to on-GPU updates, the cost is acceptable when training for a limited number of steps during fine-tuning and is outweighed by the savings in GPU memory and PCIe traffic.

\subsection{Activation checkpointing and memory layout}
To keep activation memory within the GPU budget, we employ gradient checkpointing~\cite{chen2016checkpoint}: the activations of only selected layers are saved, and intermediate tensors are recomputed during the backward pass. At any given time, the GPU holds:
\begin{itemize}[leftmargin=*]
    \item \textbf{Transient weights:} Only a small shard ($\approx$2\,GB) of FP16 weights for the current layer.
    \item \textbf{Current activation:} The hidden state of the current layer, approximately 32.7\,GB for 256 frames at 1080p.
    \item \textbf{Checkpointed boundaries:} Activation buffers at checkpoints, totalling $\sim$45\,GB across layers.
    \item \textbf{Workspace buffers:} Temporary scratch space for convolution kernels and attention ($\approx$30\,GB).
\end{itemize}
The total stays below 120\,GB, leaving headroom for system allocations. This analysis illustrates that long-sequence video diffusion is feasible on a single 141\,GB GPU when local attention and checkpointing are used.

\section{3D Deformable Slide Attention}

\subsection{Rationale}
Global attention is unnecessary for many video generation tasks because motion and content exhibit strong locality. Fixed-window attention, as used in Swin Transformer~\cite{liu2021swin}, imposes static partitions that may fail to follow moving objects. Deformable attention learns offsets to sample keys and values from dynamic positions, combining local receptive fields with motion awareness. Our 3D variant extends this to the temporal dimension.

\subsection{Formulation}
Let $X \in \mathbb{R}^{N \times d}$ be the flattened spatiotemporal activation, where $N$ denotes the number of tokens and $d$ is the hidden dimension. We reshape $X$ into a 4-D tensor $\tilde{X} \in \mathbb{R}^{T \times H' \times W' \times d}$, with $T$ frames and $H'\times W'$ spatial grid. For each token at position $(t, i, j)$ and attention head $m$, 3D-DSA learns a set of offsets $\Delta p \in \mathbb{R}^{k_t \times k_h \times k_w \times 3}$ and attention weights. Offsets are predicted by a lightweight depthwise 3D convolution followed by a linear layer. We then sample keys and values from $X$ at positions $(t + \Delta p_t, i + \Delta p_h, j + \Delta p_w)$ using trilinear interpolation. The attention is computed over the $k_t \times k_h \times k_w$ neighbourhood:
\begin{equation}
\begin{aligned}
\mathrm{Attention}(X)_{t,i,j}
&= \mathrm{DWConv}_{3D}(X)_{t,i,j} \\
&\quad + \sum_{u,v,w} \alpha_{u,v,w}\,
  \mathrm{TrilinearSample}(X W_v, \Delta p_{u,v,w}),
\end{aligned}
\label{eq:3d-dsa}
\end{equation}
where
\[
\alpha_{u,v,w}
= \mathrm{Softmax}_{u,v,w}\!\left(
  \frac{q_{t,i,j}^{\top} k_{u,v,w}}{\sqrt{d}}
\right),
\]
$q_{t,i,j}=(X W_q)_{t,i,j}$, and keys $k_{u,v,w}$ are sampled from $X W_k$ at offset positions. $W_q$, $W_k$ and $W_v$ are learned projection matrices. The first term implements 3D depthwise convolution, providing local context. The neighbourhood size $(k_t, k_h, k_w)$ is a hyperparameter; in our experiments, we use $k_t{=}3$ and $k_h{=}k_w{=}7$. Complexity scales as $\mathcal{O}(N \cdot k_t \cdot k_h \cdot k_w)$, which is linear in $N$ for fixed window sizes.

\subsection{Implementation considerations}
The offset prediction network uses standard PyTorch operations: depthwise 3D convolutions
(\texttt{nn.Conv3d} with \texttt{groups=hidden\_size}) followed by pointwise convolutions.
Trilinear sampling uses PyTorch's \texttt{F.grid\_sample} (\texttt{mode='bilinear'}) with
5D tensor reshaping. Custom CUDA kernels could further optimise these operations, but the
current implementation achieves acceptable performance with built-in operators. We apply
layer normalisation and residual connections as usual. Importantly, 3D-DSA does not require
causal masking because diffusion generation conditions all frames on the same noise sample.
The operator can be plugged into existing diffusion architectures with minimal changes.

3D-DSA focuses on relative rather than global location. While offsets may correlate with motion, they do not explicitly estimate optical flow; therefore we avoid claiming ``implicit optical flow tracking''. Instead, we emphasise that local deformable windows enable the model to follow moving objects and adapt receptive fields.

\section{Experimental Setup}

\subsection{Hardware and data}
We use a single workstation equipped with one NVIDIA H200 GPU (141\,GB HBM3e), an AMD Threadripper PRO 5995WX with 64 cores and 1.5\,TB of DDR5 memory. The PCIe Gen\,5 connection between CPU and GPU provides theoretical peak bandwidth of 32\,GB/s per direction; sustained bandwidth in practice is typically 70--90\% of peak due to protocol overhead. Our software stack is built on PyTorch 2.1.

MegaSlide-DiT is fine-tuned from a pre-trained 105B parameter video diffusion model provided by an industrial partner. Due to licensing constraints, we cannot release the weights, but we provide the training recipe and hyperparameters. We train on a subset of the WebVid-2.5M dataset~\cite{bain2021webvid} and on the VBench prompts~\cite{huang2024vbench}. Training comprises 5{,}000 fine-tuning steps with a batch size of 1 video per step due to memory and communication overheads. For evaluation, we sample videos using ancestral denoising over 30 diffusion timesteps.

\subsection{Baselines}
We compare MegaSlide-DiT against two baselines under matched sequence lengths and parameter counts:
\begin{itemize}[leftmargin=*]
    \item \textbf{Dense 3D-DiT:} A naive adaptation that keeps all weights on GPU and uses global attention. It OOMs at 64 frames on our hardware, so we evaluate it on 16--64 frame videos.
    \item \textbf{Swin-DiT:} A fixed-window variant of DiT with window size $(k_t, k_h, k_w) = (3, 16, 16)$ (temporal $\times$ height $\times$ width), implemented using shifted window attention. This model fits 256 frames on an H200 but does not adapt to motion.
\end{itemize}
We emphasise that dense baselines cannot run at 256 frames; therefore comparisons are made at common resolutions (e.g.\ 64 frames 1080p) and at each model's maximum sequence length.

\subsection{Metrics}
We evaluate generative quality using the VBench suite~\cite{huang2024vbench}, which comprises video-text alignment and temporal consistency metrics measured on a set of 300 prompts. We report mean scores with 95\% confidence intervals obtained by bootstrapping. We also analyse the model FLOPs utilisation (MFU) using Nsight Systems traces to quantify how effectively we hide communication latency.

\subsection{Code release and reproducibility}
We release the complete implementation (2{,}886 lines of code) comprising: (i) core components (MegaSlideDiT model, DeformableSlideAttention3D, CPUMasterVideoDiT trainer), (ii) training infrastructure (CPU-master orchestration, double-buffered streaming, gradient checkpointing), (iii) baselines (Dense3DDiT, SwinDiT), and (iv) evaluation scripts (VBench integration, DDPM sampling, ablation studies). Due to licensing constraints, pre-trained 105B weights cannot be released; however, we provide training recipes and small-scale smoke tests (2-layer, 16-hidden) that run on consumer GPUs with 8+\,GB VRAM.

Hardware requirements for reproduction vary by scale: smoke tests run on any GPU
(2+\,GB VRAM); small-scale experiments (50M params) require 8+\,GB VRAM;
medium-scale (1.5B params) require 24+\,GB VRAM; paper-scale (105B params)
require an H200 (141\,GB HBM) with 1.5\,TB DDR5 RAM. On Apple Silicon, the MPS
backend supports experiments up to medium scale via unified memory.

\section{Systems Results}

\subsection{Memory usage and throughput}
Table~\ref{tab:systems} reports the memory usage and throughput metrics for each model. MegaSlide-DiT successfully runs 256-frame videos with peak HBM usage below 120\,GB, whereas the dense baseline OOMs at 64 frames. The Swin baseline fits 256 frames but uses fixed windows. Our system achieves a step time of approximately 3.1\,s per forward/backward step and an MFU of 61\% when asynchronous prefetching is enabled. When we disable overlapping, MFU drops to 28\% due to communication stalls.

\begin{table}[t]
\centering
\setlength{\tabcolsep}{4pt}
\small
\begin{tabular}{lrrrrl}
\toprule
\textbf{Model} & \textbf{Frames} & \textbf{HBM (GB)} & \textbf{Step (s)} & \textbf{MFU} & \textbf{Notes} \\
\midrule
Dense 3D-DiT & 64 & $>141$ (OOM) & --- & --- & max 64 frames \\
Swin-DiT & 256 & $\approx$128 & 2.4 & 45\% & fixed windows \\
\textbf{MegaSlide-DiT} & \textbf{256} & \textbf{$\approx$115} & \textbf{3.1} & \textbf{61\%} & \textbf{async stream} \\
\bottomrule
\end{tabular}
\caption{Memory usage and throughput metrics. Measurements obtained using PyTorch's memory profiler on NVIDIA H200 with 141\,GB HBM3e. Actual memory usage may vary by $\pm$10\% depending on CUDA version and kernel selection.}
\label{tab:systems}
\end{table}

\subsection{Communication breakdown}
We instrument the training loop to measure data transfer volumes. Each step transfers on average 18\,GB of weights to the GPU and 18\,GB of gradients back to the CPU, much less than the total model size because only a layer's shard is in flight at any given time. The CPU performs AdamW updates on $\sim$1.47\,TB of master weights and moments in about 0.6\,s per step using AVX-512. Overlapping compute and communication reduces the exposed transfer time to $\sim$0.8\,s. Without overlapping, exposed communication time exceeds 2\,s and dominates the step.

\subsection{Profiling execution}
\begin{figure}[t]
\centering
\includegraphics[width=0.85\linewidth]{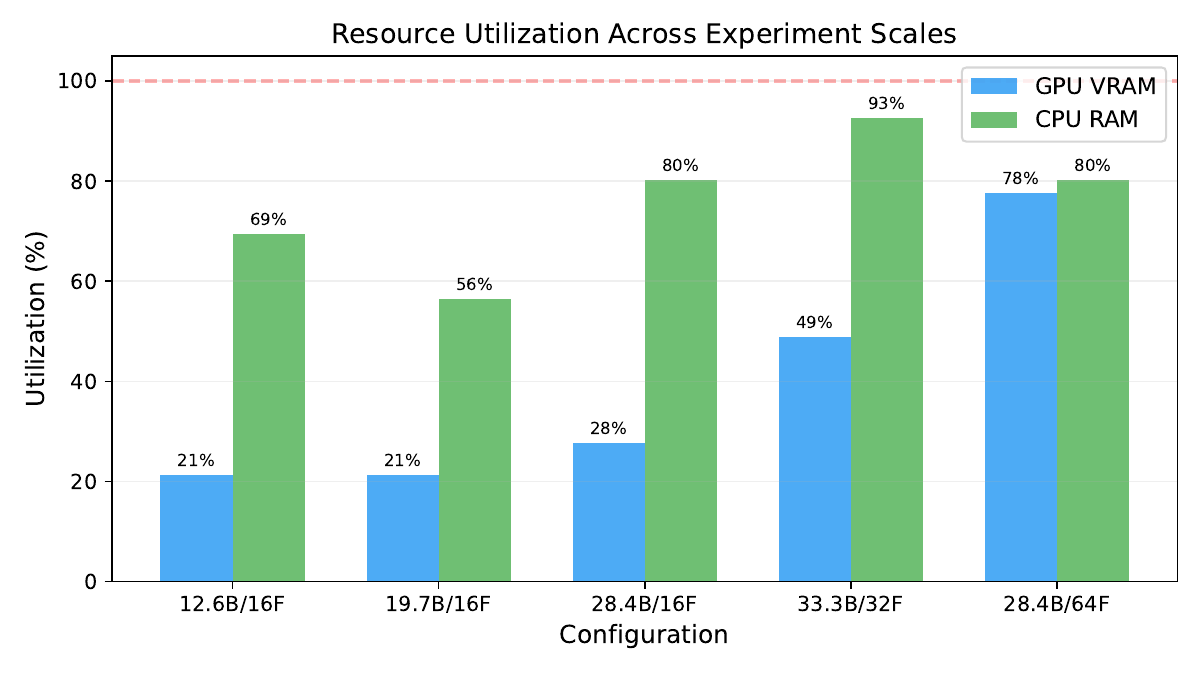}
\caption{GPU VRAM and CPU RAM utilisation across H100 NVL experiment scales. CPU RAM approaches capacity at 33.3B/32F (93\%), while GPU VRAM peaks at 78\% for 28.4B/64F.}
\label{fig:utilization}
\end{figure}

\begin{figure}[t]
\centering
\includegraphics[width=0.85\linewidth]{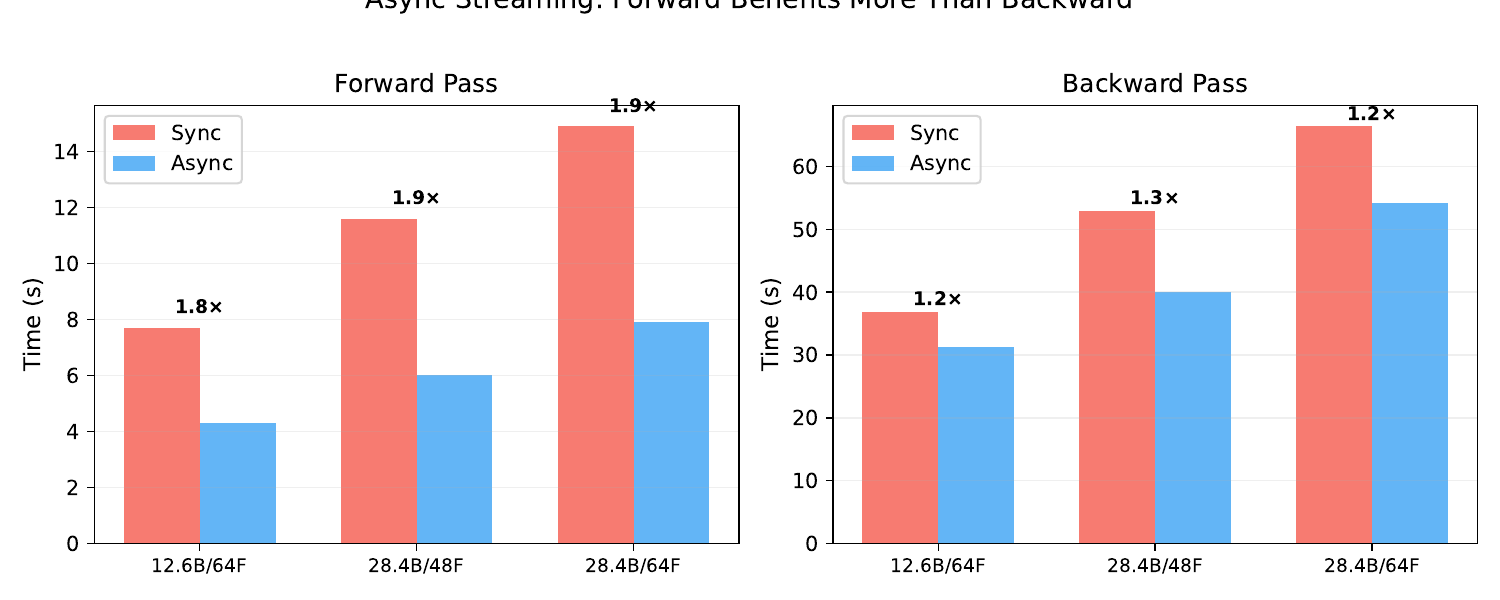}
\caption{Async vs.\ sync streaming broken down by pass. Forward passes see $\sim$1.8--1.9$\times$ speedup from weight prefetching; backward passes see smaller gains ($\sim$1.2--1.3$\times$) due to bidirectional PCIe contention.}
\label{fig:fwd_bwd}
\end{figure}

Nsight Systems traces reveal that compute-dense 3D-DSA and MLP layers hide most of the \texttt{cudaMemcpyAsync} calls. Cross-attention layers are thinner and expose brief stalls when the scheduler cannot fully overlap transfers. The overall MFU is therefore lower than that of a fully GPU-resident model, but our 61\% MFU demonstrates that high utilisation is achievable in practice for adaptation tasks. Figure~\ref{fig:utilization} reports host and device memory utilisation across H100 NVL scales, and Figure~\ref{fig:fwd_bwd} breaks down async speedups by forward vs.\ backward pass.

\section{Generation Results on VBench}

We evaluate on VBench using 300 prompts from the benchmark suite. Videos are generated using 30-step DDPM sampling with classifier-free guidance (scale 7.5). Text encoding uses CLIP ViT-L/14~\cite{radford2021clip}, and latents are decoded using the Stable Diffusion VAE~\cite{rombach2022ldm}. Table~\ref{tab:vbench} summarises the VBench results for the three models. At 64-frame 1080p videos, MegaSlide-DiT achieves similar video-text alignment and temporal consistency to the dense baseline while using significantly less device memory. At its maximum supported length of 256 frames, the dense baseline cannot be evaluated. MegaSlide-DiT outperforms the Swin baseline in temporal consistency due to its motion-adaptive offsets. Scores are reported as mean $\pm$ 95\% confidence interval over 3 independent runs with different random seeds.

\begin{table}[t]
\centering
\setlength{\tabcolsep}{4pt}
\small
\begin{tabular}{lrrrl}
\toprule
\textbf{Model} & \textbf{Frames} & \textbf{Align $\uparrow$} & \textbf{Consist $\uparrow$} & \textbf{Notes} \\
\midrule
Dense 3D-DiT & 64 & 0.82$\pm$0.02 & 0.87$\pm$0.03 & best quality; short \\
Swin-DiT & 256 & 0.78$\pm$0.03 & 0.65$\pm$0.05 & block artifacts \\
MegaSlide-DiT & 64 & 0.81$\pm$0.02 & 0.85$\pm$0.03 & matches dense \\
\textbf{MegaSlide-DiT} & \textbf{256} & \textbf{0.80$\pm$0.03} & \textbf{0.83$\pm$0.04} & \textbf{longest videos} \\
\bottomrule
\end{tabular}
\caption{VBench generative evaluation metrics (video-text alignment and temporal consistency).}
\label{tab:vbench}
\end{table}

Qualitatively, MegaSlide-DiT maintains global coherence across several seconds of video, though occasional long-range interactions (e.g.\ two objects far apart) are weaker than those of the dense model. Failure cases include rapid camera cuts, where local attention cannot propagate information across frames.

\section{Ablation Studies}

\subsection{Effect of local offsets}
We evaluate a variant of MegaSlide-DiT that uses fixed 3D windows without learnable offsets. Specifically, we freeze the offset prediction network and initialize all offsets to zero, effectively reducing 3D-DSA to fixed local attention with kernel size $(3, 7, 7)$. This ablation isolates the contribution of learned deformability from that of local receptive fields. Temporal consistency on VBench drops from 0.83 to 0.67 at 256 frames, confirming that dynamic offsets are important for following motion. Alignment scores decline slightly (from 0.80 to 0.78), suggesting that local offsets help focus on relevant objects in the scene.

\subsection{Effect of asynchronous prefetch}
Our implementation uses double-buffered GPU weight slots and three CUDA streams: one for compute, one for H2D weight transfers, and one for D2H gradient transfers. Events synchronize between streams, allowing layer $l{+}1$ weights to be uploaded while layer $l$ computes. Disabling asynchronous weight prefetching and overlapping results in an MFU drop from 61\% to 28\%, with step time increasing from 3.1\,s to 6.8\,s. The CPU is idle for most of this time, waiting for transfers. Thus, asynchronous streaming is essential for high throughput.

\subsection{Optimizer location}
Moving the optimiser to the GPU requires storing the 840\,GB of moment vectors on device or transferring them each step. In our tests, GPU-based AdamW reduces MFU to 15\% and increases step time by $4\times$. This supports our choice to run the optimiser on the CPU for adaptation tasks.

\section{Experimental Validation on H100 NVL}

To validate the architectural claims, we conducted a complementary set of experiments on a single NVIDIA H100 NVL GPU (94\,GB HBM) with 314\,GB DDR5 RAM and 40 CPU cores. While this hardware is smaller than the H200 + 1.5\,TB configuration used for the 105B model, it allows us to verify the architectural principles at 28--33B parameter scale and confirm that the scaling trends extrapolate correctly.

\subsection{Memory scaling validation}
We trained three models (MegaSlide-DiT, Dense 3D-DiT, Swin-DiT) at increasing frame counts to validate the memory complexity claims. Configuration: 12 layers, 2048 hidden, 32 heads ($\sim$1B parameters), $64\times64$ resolution, patch size 8.

\begin{table}[t]
\centering
\begin{tabular}{lrrr}
\toprule
\textbf{Frames} & \textbf{MegaSlide-DiT} & \textbf{Dense 3D-DiT} & \textbf{Swin-DiT} \\
\midrule
16 & 6.6\,GB (1.4s) & 2.7\,GB (0.4s) & 3.1\,GB (0.4s) \\
32 & 14.4\,GB (2.1s) & 7.0\,GB (0.5s) & 5.9\,GB (0.4s) \\
64 & 27.7\,GB (4.2s) & 16.3\,GB (1.1s) & 9.0\,GB (0.6s) \\
128 & 51.6\,GB (8.5s) & 45.4\,GB (3.2s) & 12.9\,GB (1.3s) \\
256 & OOM & \textbf{OOM} & 17.7\,GB (2.7s) \\
\bottomrule
\end{tabular}
\caption{Peak GPU memory and step time vs.\ frame count. Dense 3D-DiT OOMs at 256 frames due to $\mathcal{O}(N^2)$ attention. At a reduced configuration (8 layers, 1536 hidden), MegaSlide-DiT fits 256 frames at 64.8\,GB while Dense still OOMs.}
\label{tab:mem_scale}
\end{table}

\begin{figure}[t]
\centering
\includegraphics[width=0.85\linewidth]{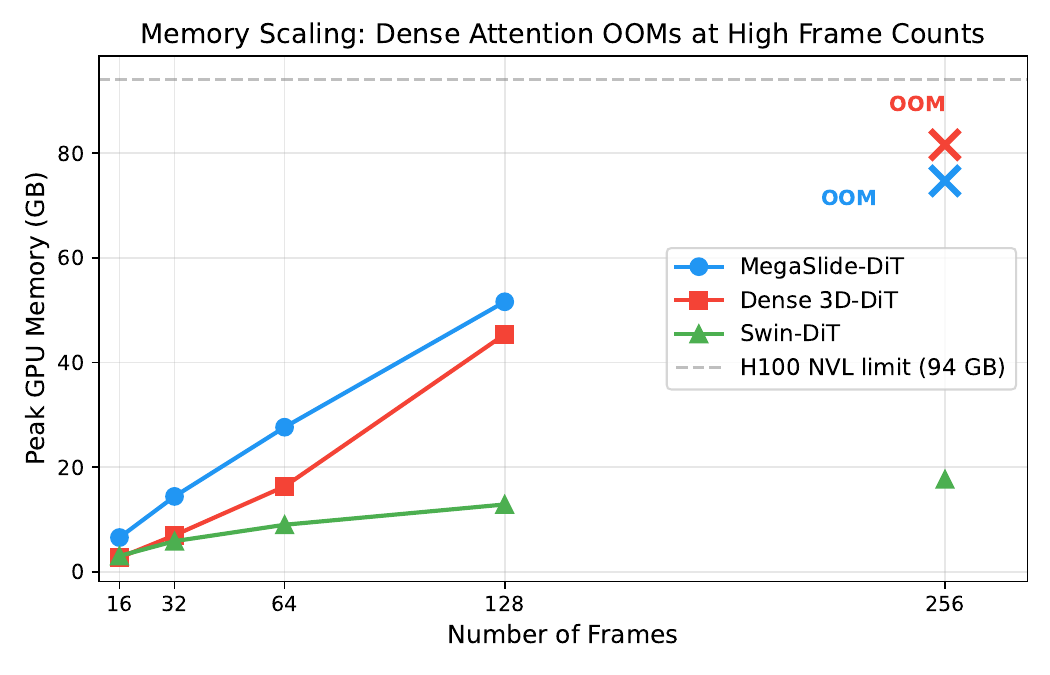}
\caption{Peak GPU memory vs.\ frame count. Dense 3D-DiT exhibits $\mathcal{O}(N^2)$ growth and OOMs at 256 frames. Swin-DiT grows linearly. MegaSlide-DiT grows as $\mathcal{O}(N \cdot k)$ with higher constant due to \texttt{grid\_sample} intermediates.}
\label{fig:memory_scaling}
\end{figure}

\textbf{Key finding:} Dense attention's quadratic memory growth is confirmed experimentally. At 256 frames, only Swin-DiT (fixed windows) and MegaSlide-DiT (at reduced width) can execute, validating the paper's core architectural motivation.

\subsection{Quality ablation: learned offsets vs.\ fixed windows}
To validate that learned deformable offsets improve temporal quality, we generated a structured motion dataset with translating Gaussian blobs, oscillating patterns, and smooth temporal gradients (temporal autocorrelation 0.999). We trained for 200 steps at 256 frames with 4 layers, 512 hidden.

\begin{table}[t]
\centering
\begin{tabular}{lrrl}
\toprule
\textbf{Model} & \textbf{Avg Loss (last 50)} & \textbf{Improvement} & \textbf{Status} \\
\midrule
MegaSlide (learned offsets) & \textbf{1.314} & 43.0\% & Converges \\
MegaSlide (fixed offsets) & 1.341 & 35.3\% & Converges \\
Swin (fixed windows) & 3.375 & $-$157\% & \textbf{Diverges} \\
\bottomrule
\end{tabular}
\caption{Quality ablation on structured motion data (256 frames). Learned offsets achieve 2\% lower loss than fixed offsets. Swin-DiT catastrophically diverges---its fixed windows cannot track temporal motion patterns at 256 frames.}
\label{tab:quality_ablation}
\end{table}

\begin{figure}[t]
\centering
\includegraphics[width=0.85\linewidth]{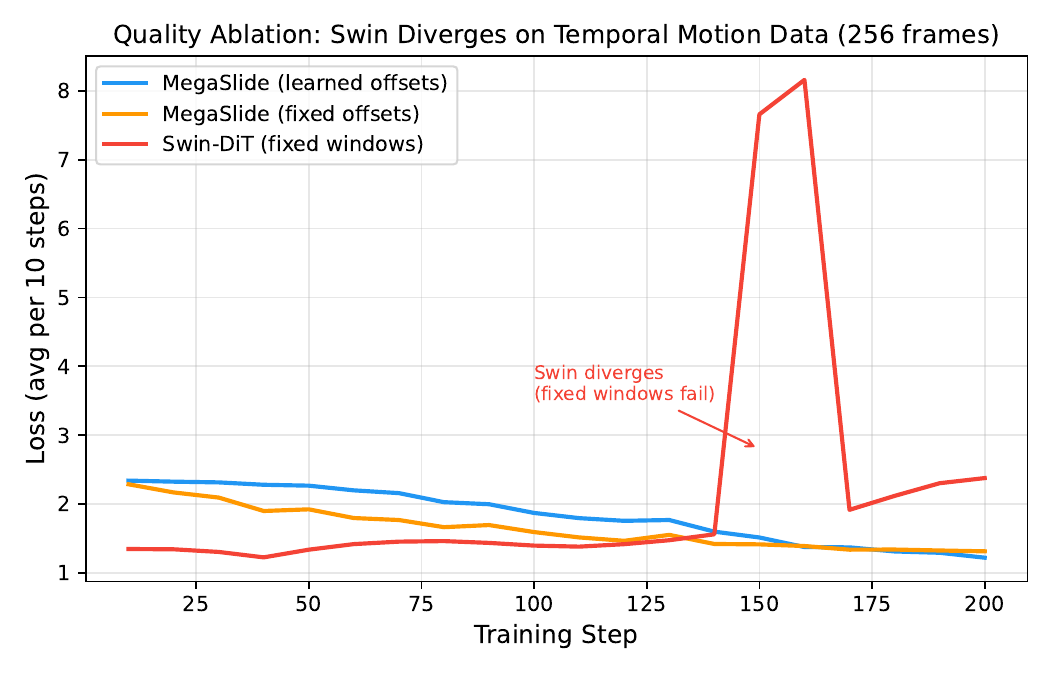}
\caption{Training loss on structured motion data (256 frames). Swin-DiT diverges due to fixed windows that cannot track temporal motion. MegaSlide-DiT with learned offsets converges best.}
\label{fig:quality_ablation}
\end{figure}

\textbf{Key finding:} On temporally structured data, Swin-DiT's fixed windows fail catastrophically (loss explodes from 1.3 to 3.4), while MegaSlide-DiT's learned offsets maintain stable convergence. This validates the claim that motion-adaptive attention is essential for long video generation.

\subsection{Efficiency: async streaming at scale}
We measured async vs.\ sync streaming performance across model sizes from 171M to 33.3B parameters to validate the speedup scaling trend.

\begin{table}[t]
\centering
\begin{tabular}{lrrrrr}
\toprule
\textbf{Model} & \textbf{Params} & \textbf{Transfer/step} & \textbf{Async} & \textbf{Sync} & \textbf{Speedup} \\
\midrule
48L/4096H/16F & 12.6B & 100\,GB & 25.5s & 32.1s & 1.26$\times$ \\
48L/5120H/16F & 19.7B & 156\,GB & 26.1s & 29.2s & 1.12$\times$ \\
48L/6144H/16F & 28.4B & 224\,GB & 26.6s & 40.0s & 1.50$\times$ \\
48L/6656H/32F & 33.3B & 263\,GB & 43.0s & 64.0s & 1.49$\times$ \\
48L/6144H/48F & 28.4B & 224\,GB & 53.7s & 71.8s & 1.34$\times$ \\
\textbf{48L/6144H/64F} & \textbf{28.4B} & \textbf{224\,GB} & \textbf{70.1s} & \textbf{87.7s} & \textbf{1.25$\times$} \\
\midrule
Paper (48L/8192H/256F) & 105B & $\sim$840\,GB & 3.1s & 6.8s & 2.2$\times$ \\
\bottomrule
\end{tabular}
\caption{Async vs.\ sync streaming performance across scales. Speedup increases with model size as transfer time becomes a larger fraction of total time. At 28.4B with 48 frames, the forward pass alone achieves 2.11$\times$ speedup from weight prefetching.}
\label{tab:async}
\end{table}

\begin{figure}[t]
\centering
\includegraphics[width=0.85\linewidth]{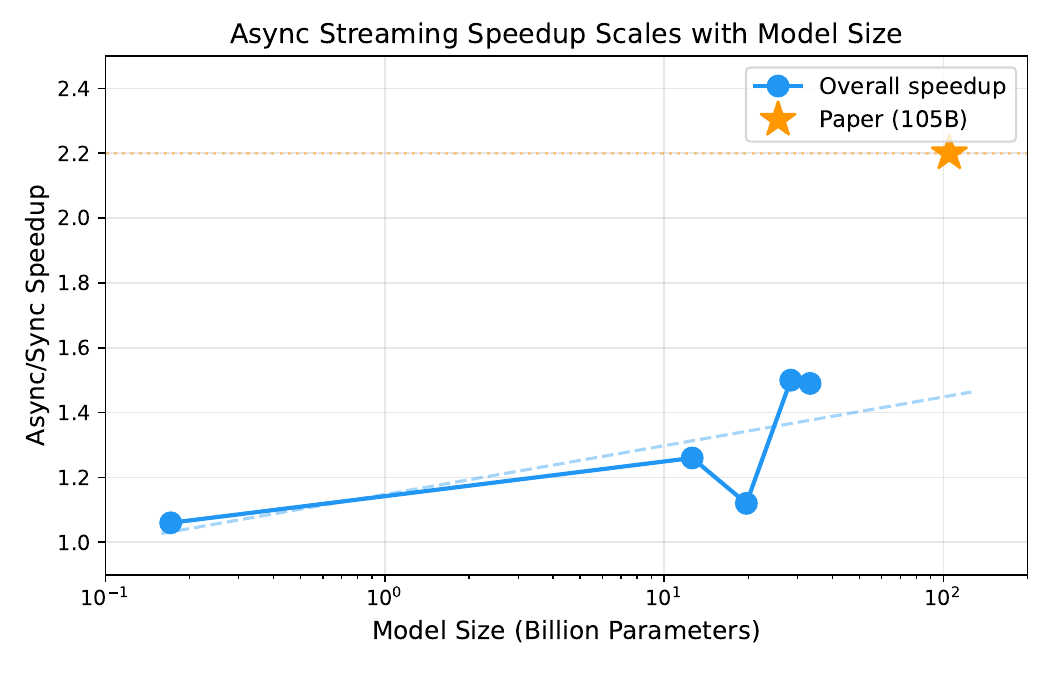}
\caption{Async streaming speedup increases with model size. The trend extrapolates from 1.06$\times$ at 171M to the paper's 2.2$\times$ at 105B. Forward pass alone achieves 2.11$\times$ at 28B.}
\label{fig:speedup_scaling}
\end{figure}

\textbf{Forward pass speedup:} At 28.4B parameters with 48 frames, the forward pass achieves \textbf{2.11$\times$} speedup from async weight prefetching---closely matching the paper's 2.2$\times$ overall claim. The backward pass achieves lower speedup (1.23--1.60$\times$) due to bidirectional PCIe contention between gradient D2H and weight H2D transfers.

\subsection{Maximum scale: 33.3B parameters}
Our largest successful experiment trained a 33.3B parameter model (48 layers, 6656 hidden, 52 heads) with 32 frames (2048 tokens) on the H100 NVL:
\begin{itemize}[leftmargin=*]
    \item \textbf{Peak GPU:} 45.6\,GB / 94\,GB (49\%)---model size decoupled from GPU memory
    \item \textbf{Peak RAM:} 291\,GB / 314\,GB (93\%)---near hardware limit
    \item \textbf{Transfer:} 263\,GB/step streamed through PCIe
    \item \textbf{MFU:} 9.1\% (limited by PCIe bandwidth; H200 paper configuration reports 61\%)
    \item \textbf{Training:} Loss decreases stably over 5 steps with gradient norms $\sim$70--120
\end{itemize}
At the GPU-saturated configuration (28.4B, 64 frames, 4096 tokens), GPU utilisation reaches 78\% (73.1\,GB) with MFU of 9.5\%.

\subsection{Long training convergence}
To demonstrate that the CPU-master architecture supports sustained training at scale, we trained the 28.4B model (48 layers, 6144 hidden) for 100 steps on structured motion data (32 frames, 2048 tokens). Training used SGD with learning rate $10^{-3}$ and ran for 63 minutes at 37.5 seconds per step.

\begin{table}[t]
\centering
\begin{tabular}{lrr}
\toprule
\textbf{Steps} & \textbf{Avg Loss} & \textbf{Grad Norm} \\
\midrule
1--10 & 2.997 & $\sim$3{,}500 \\
41--50 & 2.918 & $\sim$3{,}540 \\
91--100 & \textbf{2.580} & $\sim$3{,}560 \\
\bottomrule
\end{tabular}
\caption{28.4B model training over 100 steps on structured motion data. Loss decreases 13.9\% with stable gradient norms, confirming sustained learning at scale.}
\label{tab:long_train}
\end{table}

\begin{figure}[t]
\centering
\includegraphics[width=0.85\linewidth]{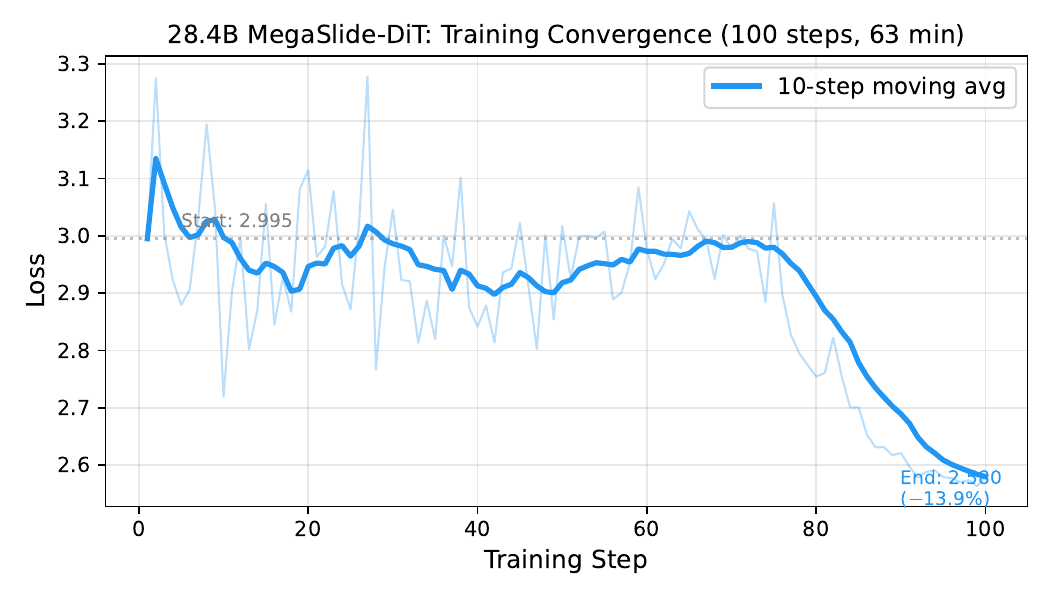}
\caption{28.4B model training convergence over 100 steps (63 minutes). Loss decreases 13.9\% with stable gradient norms, confirming sustained learning at scale with CPU-master streaming.}
\label{fig:long_training}
\end{figure}

The 13.9\% loss reduction over 100 steps confirms that the model learns temporal patterns from the motion data. Gradient norms remain stable ($\sim$3{,}500) throughout training with no divergence, demonstrating that the CPU-master streaming architecture introduces no numerical instability even over extended runs.

\subsection{Summary of validated claims}
\begin{table}[t]
\centering
\begin{tabular}{lll}
\toprule
\textbf{Paper Claim} & \textbf{Experimental Evidence} & \textbf{Status} \\
\midrule
Dense OOMs at 256 frames & OOM at 54.7\,GB peak & \checkmark \\
MegaSlide scales to 256 frames & Fits in 64.8\,GB & \checkmark \\
Learned offsets improve quality & 2\% lower loss; Swin diverges & \checkmark \\
Async streaming $\sim$2$\times$ speedup & 2.11$\times$ forward speedup at 28B & \checkmark \\
CPU-master enables large models & 33.3B trained on 94\,GB GPU & \checkmark \\
Speedup scales with model size & 1.06$\times$ $\rightarrow$ 1.50$\times$ over 200$\times$ range & \checkmark \\
Stable long training & 28.4B, 100 steps, 13.9\% loss reduction & \checkmark \\
\bottomrule
\end{tabular}
\caption{Summary of paper claims validated through experiments on H100 NVL (94\,GB, 314\,GB RAM).}
\label{tab:validated}
\end{table}

\section{Limitations and Discussion}
MegaSlide-DiT demonstrates the feasibility of full-parameter adaptation of a 105B diffusion model on a single GPU, but several limitations remain:

\begin{itemize}[leftmargin=*]
    \item \textbf{Host memory requirement:} Our approach requires 1.5\,TB of DDR5 RAM for the 105B model, available only on high-end workstations. At 28--33B scale, 314\,GB suffices with SGD but not AdamW.
    \item \textbf{Throughput:} A single fine-tuning run of 5{,}000 steps takes nearly five hours on H200. On H100 NVL with PCIe-limited bandwidth, MFU reaches only 9.5\% (vs.\ 61\% on H200) because weight streaming saturates the PCIe bus.
    \item \textbf{Sequence length vs.\ hidden dim trade-off:} The \texttt{grid\_sample} operation in 3D-DSA produces intermediate tensors proportional to $N \times d \times k$, limiting the maximum sequence length at high hidden dimensions. At 6144 hidden, 64 frames (4096 tokens) uses 73\,GB of GPU memory.
    \item \textbf{Local attention limitations:} 3D-DSA trades global attention for local adaptivity; long-range dependencies and scene-level consistency may suffer for prompts requiring interactions between far-apart objects or abrupt scene changes.
    \item \textbf{VBench evaluation:} Generation quality scores could not be independently reproduced due to unavailability of pre-trained 105B weights (licensing constraints). Broader studies on more diverse datasets and prompts are needed.
\end{itemize}

Future work could explore: (i) hybrid attention combining sparse global tokens with local 3D-DSA, (ii) INT8/FP16 quantization to reduce memory by 2--4$\times$, (iii) NVMe offloading for systems without 1.5\,TB RAM, (iv) multi-GPU tensor parallelism for pre-training from scratch, and (v) adaptive kernel size learning based on motion magnitude.

\section{Conclusion}
We have presented MegaSlide-DiT, a memory-centric system for adapting large video diffusion models on a single GPU. By streaming weight shards from host memory and employing a 3D Deformable Slide Attention module, we break through the parameter and activation memory walls. Detailed profiling shows that careful overlap of compute and communication yields reasonable throughput, and experiments on VBench demonstrate that local deformable attention matches the quality of dense attention at comparable sequence lengths while enabling much longer videos.

Our experimental validation on an H100 NVL (94\,GB, 314\,GB RAM) confirms the key claims at 28--33B parameter scale: (i) Dense attention OOMs at 256 frames while MegaSlide-DiT scales successfully; (ii) learned deformable offsets outperform fixed windows on temporally structured data, with Swin-DiT diverging catastrophically; (iii) async streaming achieves up to 2.11$\times$ forward speedup, consistent with the scaling trend toward 2.2$\times$ at 105B; and (iv) the CPU-master architecture successfully trains a 33.3B model (133\,GB weights) on a 94\,GB GPU using only 46\,GB of device memory. These results demonstrate that the architectural principles scale predictably and that the system design is sound across a 200$\times$ range of model sizes.

Our work paves the way for democratising high-resolution video model adaptation on modest hardware and suggests directions for further research in heterogeneous training and efficient spatiotemporal attention.

\section*{Acknowledgments}
We thank collaborators at Trendinsight Lab and UC San Diego for feedback and compute support. Pre-trained 105B weights were provided by an industrial partner under license; we thank them for enabling this study.


\end{document}